# EXPERIMENT ON CREATING A NEURAL NETWORK WITH WEIGHTS DETERMINED BY THE POTENTIAL OF A SIMULATED ELECTROSTATIC FIELD

**P.Sh. Geidarov**
**Institute of Control Systems, Baku, Azerbaijan**
plbaku2010@gmail.com

**Abstract.**This paper explores the possibility of determining the weights and thresholds of a neural network using the potential—a parameter of an electrostatic field—without analytical calculations and without applying training algorithms. The work is based on neural network architectures employing metric recognition methods. The electrostatic field is simulated in the Builder C++ environment. In the same environment, a neural network based on metric recognition methods is constructed, with the weights of the first-layer neurons determined by the values of the potentials of the simulated electrostatic field. The effectiveness of the resulting neural network within the simulated system is evaluated using the MNIST test dataset under various initial conditions of the simulated system. The results demonstrated functional viability. The implementation of this approach shows that a neural network can obtain weight values almost instantaneously from the electrostatic field, without the need for analytical computations, lengthy training procedures, or massive training datasets.

**Introduction and Purpose of the Work.**

The use of artificial neural networks remains a promising direction in automated systems, particularly in the context of pattern recognition tasks. Although artificial neural networks were originally conceived as systems that mimic biological neural networks [1,2], the capabilities of modern classical artificial neural networks still fall short of their biological counterparts by many criteria. These include, among others, such abilities of the biological brain as the capacity to store a vast number of recognizable patterns, as well as the ability to rapidly memorize and recognize new objects and patterns without the need for prolonged training procedures and massive training datasets. To address this, studies [3,4,5] have explored neural network architectures based on metric recognition methods [6]. These networks possess a number of distinctive capabilities, including the following:

1. The ability to quickly create the structure (number of layers, neurons, and connections) of a neural network based on the initial parameters of the task, such as the number of templates and patterns.
2. The ability to easily and simply expand the neural network by sequentially adding new patterns without the need for retraining the network or modifying the previously obtained weight values.
3. The ability to pre-calculate the neural network weight values based on the expressions of metric proximity measures [6].
4. The ability to further train the resulting neural network using classical training algorithms.

It should be noted that the pre-calculation of weight values helps accelerate the process of creating and training the neural network. However, analytical computation of weight values also requires time, which increases with the number of recognizable patterns, the number of templates used in the task, and the dimensionality of the weight table. The latter is also related to image quality (such as pixel dimensions, etc.). Furthermore, in the biological world of wildlife, there are no mechanisms, processors, or other computational devices that could perform analytical calculations. In this regard, the work [7] demonstrated the possibility of determining the weight values of a neural network based on the parameters of an electrostatic field, which essentially means the possibility of almost instantaneous determination of the network's weights without using analytical calculations.

The aim of this work is the practical verification of the operability of a neural network in which the weights are defined based on an electrostatic field parameter — the potential. To achieve this, a neural network is implemented in the Builder C++ programming environment, using a simulated electrostatic field, where the weights are determined based on the electrostatic field potential values. The operability of the resulting neural network is also tested on the handwritten digit recognition task using the MNIST benchmark dataset.

## 1. Creation of a Neural Network and Calculation of Weight Potentials

Let us recall that the MNIST dataset consists of 10,000 images of handwritten digits, each with a resolution of 28 by 28 pixels (Fig. 1). Each pixel of the image has a grayscale value in the range [0, 255]. In this example, each image in the MNIST dataset is considered as a monochrome black-and-white image. The grayscale values of MNIST images are interpreted as either black or white depending on the pixel intensity: values > 150 are treated as white, and values < 150 as black. Each white pixel is assumed to represent a region with a point charge of $q = 10^{-9}$ coulombs. Black pixels are considered as uncharged regions of the image. The distance between adjacent cells in a single image is set to $d_1$ = 2 cm (see Fig. 1). Throughout the text, the designation of MNIST images will follow the format of the digit class label followed by the image's index in the dataset, for example: 0_157, 1_46 (Fig. 1).

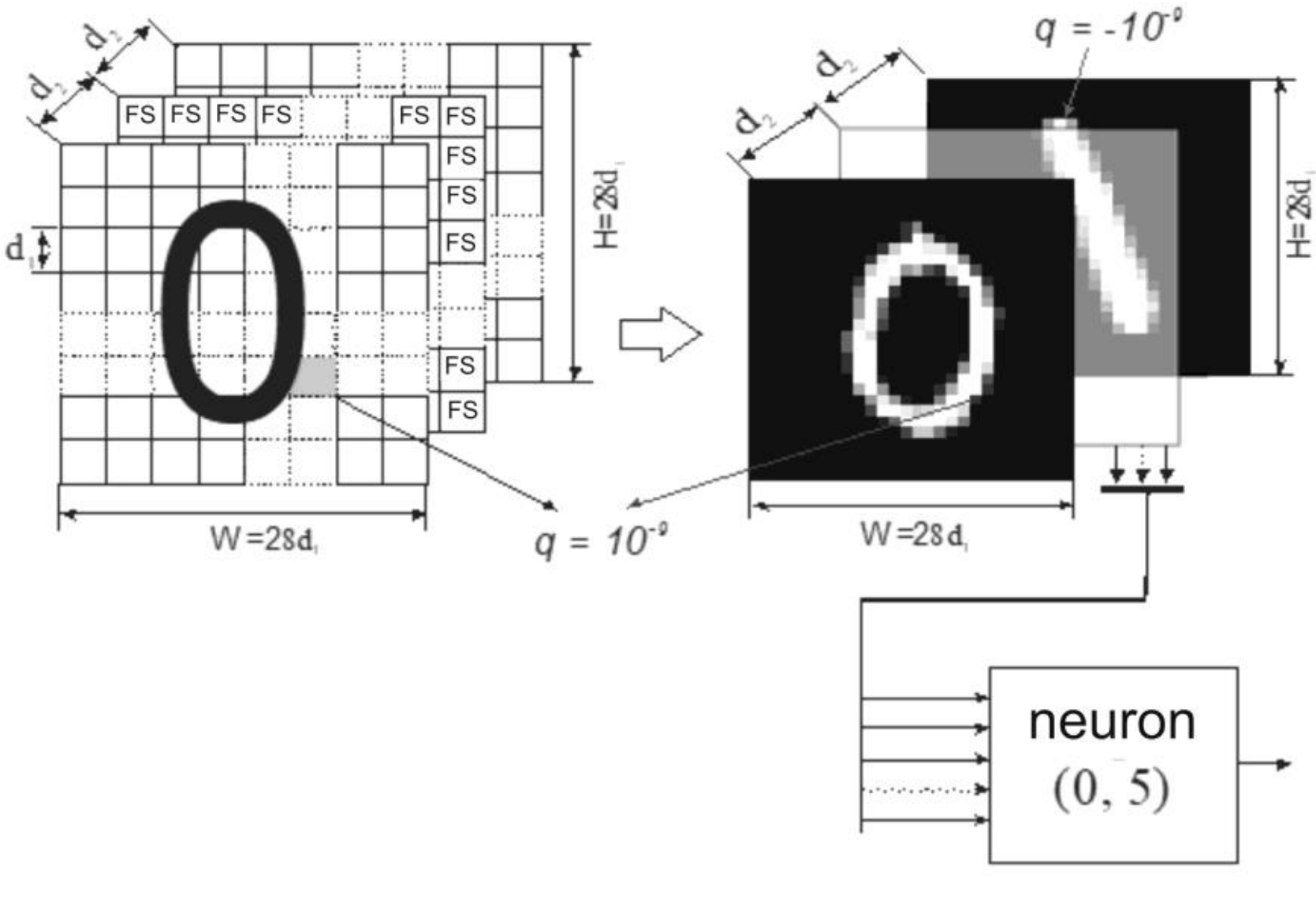

Fig. 1. Diagram of the simulated model of charged planes representing images 0_157 and 1_46 from the MNIST dataset, and the plane of potentiometric sensors (PS), which defines a neuron of the first layer of the neural network.

The distance between the two compared images in this example is set to 8 cm. That is, the plane of the simulated potentiometric sensors is located between the two planes of the charged image surfaces at a distance of $\mathbf{d_2 = 4\ cm}$ from each image (see Fig. 1). It should be noted that the simulated system can be considered both for a neural network based on metric recognition methods with a zero layer [4], and for a neural network without a zero layer (see Fig. 2).

In the case with a zero layer, and unlike the scheme in Fig. 2, each reference image corresponds to its own plane of potentiometric sensors, in which the potential values are determined for that reference image only. Since in this case the number of weight tables is reduced and becomes equal to the number of reference images used [4], less measuring equipment is required accordingly. It is also possible to use a single potentiometric plane with sequential application and determination of weight tables for each neuron of the zero or first layer.

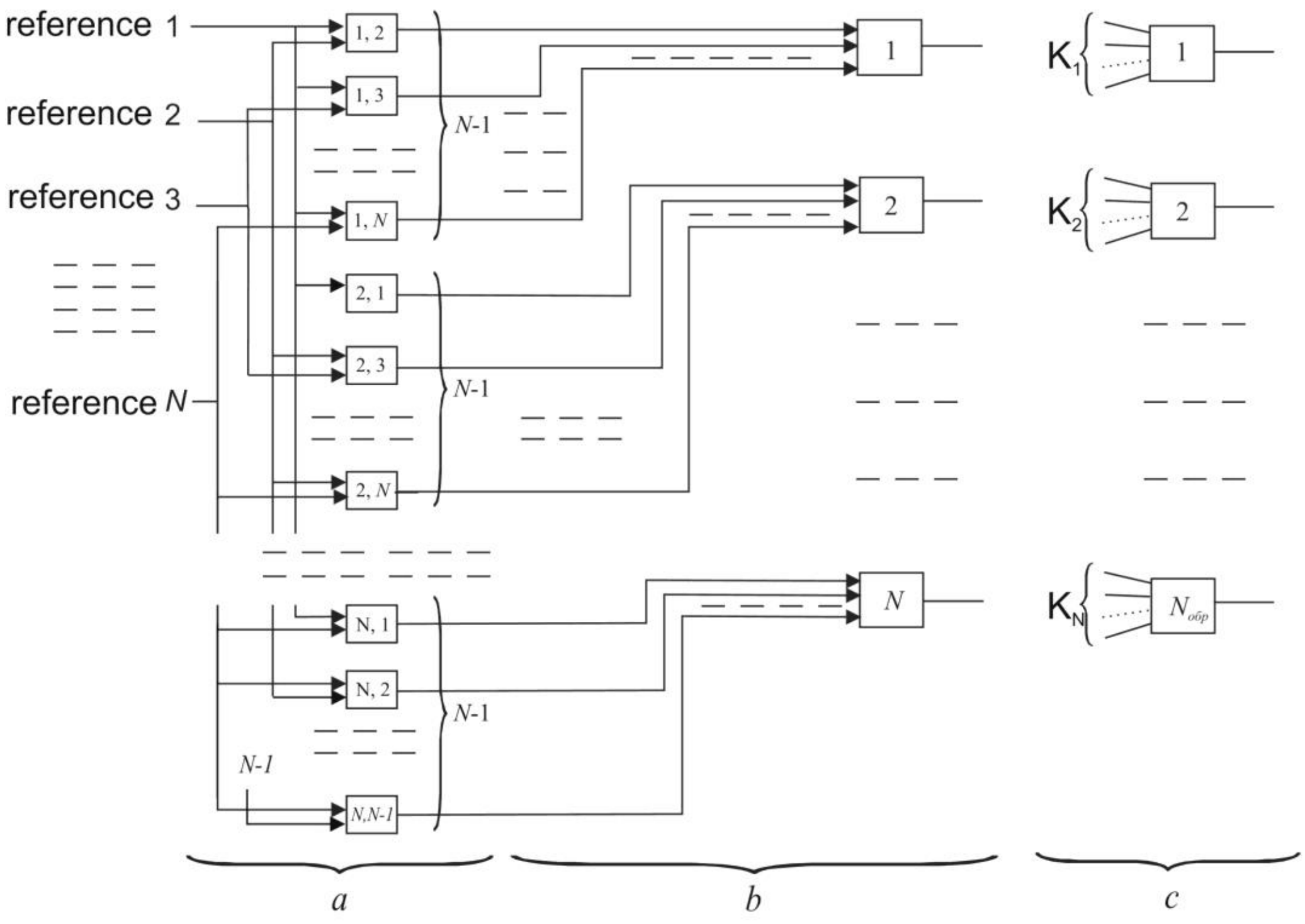

Fig. 2. Extended diagram of a neural network based on metric recognition methods without a zero layer.

The value of each weight in the weight table of the first layer $w_{(i,j)}^{(1)} = \varphi_{(i,j)}^{Sum}$ (Fig. 3) is defined as the sum of the potentials of point charges located in the active pixels (i,j)(i, j)(i,j) of two images (1).

$$\varphi_{(i,j)}^{Sum} = \sum_{\substack{i=0,\\ j=0}}^{\substack{i=28,\\ j=28}} \varphi_{(i,j)}^{(Img1)} - \sum_{\substack{i=0,\\ j=0}}^{\substack{i=28,\\ j=28}} \varphi_{(i,j)}^{(Img2)} = \sum_{\substack{i=0,\\ j=0}}^{\substack{i=28,\\ j=28}} \frac{K*q_{(i,j)}^{(Img1)}}{r_{(i,j)}^{(Img1)}} - \sum_{\substack{i=0,\\ j=0}}^{\substack{i=28,\\ j=28}} \frac{K*q_{(i,j)}^{(Img2)}}{r_{(i,j)}^{(Img1)}}, \quad (1)$$

where $\varphi_{(i,j)}^{(Img1)}$ , $\varphi_{(i,j)}^{(Img2)}$ are the total potentials from the first image plane Img1 and the second image plane Img2 (Figs. 1, 3).

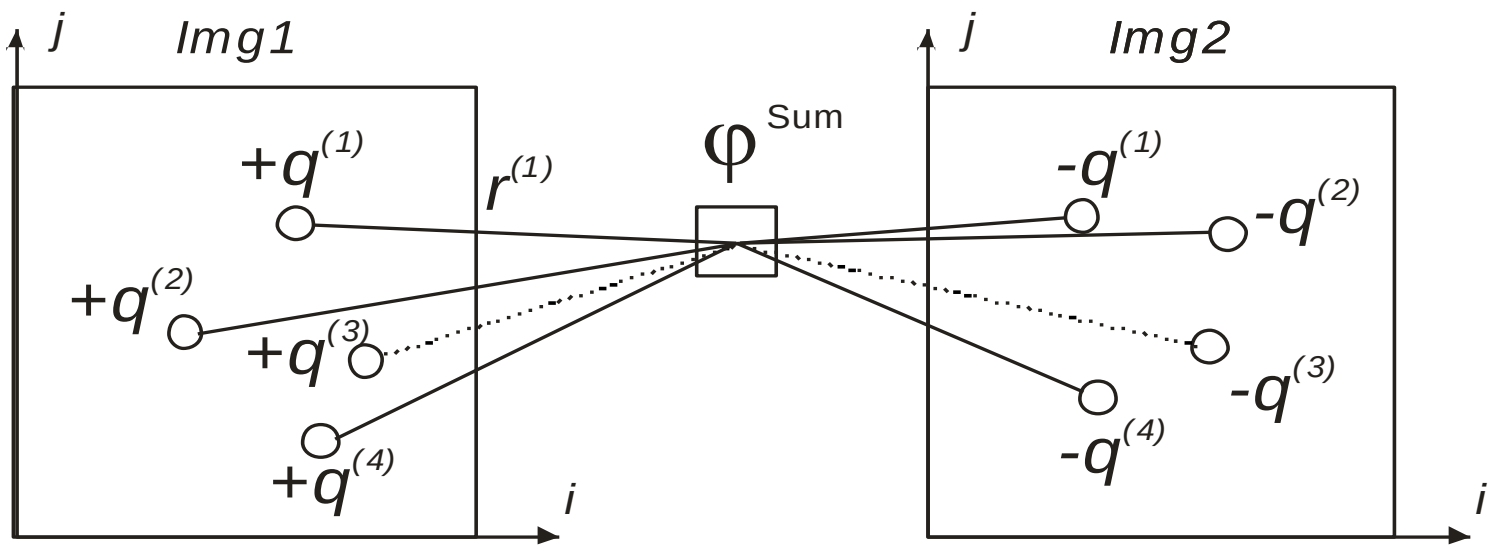

Fig. 3. Diagram of determining the potential-weight in a single cell of the weight table (sensor plane of potentiometers) of the first layer.

Table 1. Weight table of neuron 0 of the zero layer for image 0_157 (Fig. 1)

| k = 0, 0_157 | | | | | | | | | | | | | | | | | | | | | | | | | | | |
|---|---|---|---|---|---|---|---|---|---|---|---|---|---|---|---|---|---|---|---|---|---|---|---|---|---|---|---|
| 927 | 012 | 101 | 190 | 284 | 388 | 489 | 595 | 702 | 806 | 904 | 989 | 061 | 111 | 135 | 126 | 084 | 020 | 937 | 844 | 742 | 638 | 530 | 425 | 327 | 228 | 136 | 051 |
| 996 | 088 | 187 | 291 | 401 | 517 | 642 | 769 | 901 | 033 | 159 | 276 | 374 | 443 | 479 | 465 | 407 | 318 | 206 | 079 | 949 | 818 | 689 | 565 | 446 | 336 | 231 | 132 |
| 066 | 169 | 277 | 395 | 523 | 658 | 806 | 963 | 124 | 295 | 462 | 622 | 765 | 866 | 919 | 902 | 816 | 688 | 526 | 355 | 186 | 017 | 861 | 714 | 574 | 446 | 329 | 219 |
| 138 | 248 | 369 | 503 | 649 | 806 | 982 | 174 | 379 | 599 | 825 | 050 | 260 | 426 | 518 | 493 | 353 | 149 | 913 | 674 | 451 | 241 | 048 | 871 | 709 | 563 | 426 | 302 |
| 207 | 328 | 465 | 612 | 776 | 961 | 172 | 404 | 663 | 948 | 262 | 590 | 909 | 188 | 380 | 342 | 079 | 740 | 386 | 047 | 746 | 484 | 249 | 036 | 849 | 680 | 529 | 388 |
| 276 | 407 | 554 | 721 | 909 | 123 | 368 | 649 | 973 | 350 | 786 | 272 | 765 | 177 | 500 | 442 | 011 | 504 | 953 | 470 | 071 | 739 | 458 | 210 | 993 | 799 | 629 | 475 |
| 337 | 483 | 644 | 828 | 039 | 283 | 568 | 907 | 309 | 803 | 408 | 102 | 771 | 174 | 369 | 280 | 920 | 356 | 574 | 918 | 414 | 012 | 677 | 389 | 139 | 920 | 727 | 562 |
| 397 | 555 | 730 | 929 | 164 | 439 | 766 | 162 | 659 | 302 | 111 | 936 | 449 | 639 | 676 | 518 | 259 | 849 | 120 | 388 | 784 | 300 | 902 | 568 | 283 | 040 | 825 | 641 |
| 452 | 618 | 808 | 028 | 285 | 587 | 954 | 410 | 995 | 787 | 766 | 452 | 655 | 497 | 408 | 155 | 993 | 928 | 576 | 894 | 186 | 601 | 133 | 749 | 426 | 152 | 921 | 720 |
| 498 | 676 | 879 | 113 | 389 | 720 | 126 | 637 | 291 | 150 | 111 | 584 | 502 | 122 | 881 | 693 | 611 | 675 | 715 | 354 | 600 | 912 | 366 | 922 | 560 | 259 | 008 | 791 |
| 542 | 727 | 939 | 188 | 478 | 835 | 279 | 839 | 565 | 451 | 231 | 402 | 083 | 716 | 480 | 350 | 327 | 423 | 598 | 601 | 998 | 233 | 591 | 086 | 681 | 357 | 083 | 852 |
| 573 | 765 | 987 | 248 | 553 | 930 | 404 | 014 | 828 | 775 | 261 | 075 | 708 | 404 | 220 | 134 | 147 | 286 | 563 | 813 | 426 | 555 | 798 | 230 | 787 | 436 | 148 | 904 |
| 596 | 791 | 024 | 289 | 605 | 999 | 493 | 135 | 997 | 941 | 261 | 889 | 482 | 211 | 057 | 003 | 053 | 234 | 590 | 969 | 683 | 781 | 954 | 338 | 871 | 498 | 196 | 943 |
| 611 | 808 | 041 | 313 | 639 | 040 | 544 | 200 | 070 | 993 | 246 | 803 | 368 | 102 | 966 | 934 | 010 | 222 | 625 | 059 | 812 | 900 | 049 | 413 | 926 | 537 | 227 | 968 |
| 613 | 813 | 046 | 319 | 648 | 054 | 559 | 219 | 085 | 995 | 220 | 752 | 308 | 046 | 923 | 910 | 003 | 237 | 657 | 103 | 864 | 950 | 089 | 444 | 950 | 555 | 241 | 978 |
| 605 | 805 | 037 | 306 | 630 | 034 | 540 | 194 | 056 | 963 | 183 | 712 | 271 | 023 | 914 | 917 | 030 | 273 | 690 | 118 | 861 | 940 | 075 | 431 | 939 | 549 | 235 | 972 |
| 588 | 785 | 012 | 278 | 596 | 989 | 484 | 130 | 987 | 892 | 126 | 677 | 261 | 033 | 946 | 969 | 101 | 354 | 747 | 112 | 795 | 858 | 008 | 378 | 898 | 516 | 212 | 953 |
| 561 | 750 | 973 | 230 | 534 | 912 | 393 | 019 | 860 | 771 | 034 | 639 | 273 | 081 | 027 | 080 | 243 | 532 | 889 | 074 | 600 | 670 | 876 | 283 | 823 | 463 | 168 | 917 |
| 525 | 710 | 918 | 162 | 453 | 811 | 262 | 850 | 652 | 558 | 880 | 600 | 336 | 199 | 172 | 258 | 482 | 843 | 151 | 976 | 223 | 377 | 686 | 151 | 727 | 389 | 109 | 868 |
| 479 | 654 | 853 | 081 | 354 | 685 | 090 | 619 | 325 | 147 | 589 | 600 | 522 | 427 | 418 | 522 | 790 | 192 | 240 | 732 | 844 | 069 | 471 | 993 | 611 | 297 | 034 | 809 |
| 428 | 592 | 779 | 989 | 237 | 534 | 893 | 335 | 885 | 511 | 091 | 604 | 816 | 778 | 803 | 902 | 090 | 305 | 039 | 249 | 421 | 757 | 240 | 819 | 477 | 193 | 950 | 740 |
| 372 | 526 | 692 | 884 | 113 | 372 | 680 | 043 | 470 | 967 | 562 | 257 | 728 | 015 | 202 | 293 | 290 | 129 | 586 | 728 | 997 | 440 | 998 | 636 | 333 | 076 | 854 | 664 |
| 310 | 451 | 605 | 778 | 981 | 209 | 470 | 770 | 111 | 509 | 987 | 532 | 072 | 642 | 020 | 099 | 911 | 514 | 843 | 149 | 578 | 126 | 759 | 448 | 184 | 954 | 754 | 581 |
| 246 | 373 | 512 | 670 | 845 | 046 | 268 | 517 | 793 | 108 | 463 | 857 | 296 | 794 | 238 | 298 | 976 | 566 | 075 | 595 | 177 | 824 | 522 | 260 | 031 | 830 | 652 | 492 |
| 179 | 292 | 420 | 563 | 716 | 885 | 076 | 283 | 508 | 752 | 021 | 311 | 615 | 932 | 190 | 224 | 030 | 750 | 436 | 114 | 810 | 540 | 297 | 076 | 879 | 703 | 548 | 407 |
| 108 | 213 | 326 | 453 | 589 | 734 | 894 | 068 | 250 | 442 | 646 | 856 | 064 | 248 | 378 | 398 | 302 | 136 | 928 | 708 | 486 | 276 | 082 | 898 | 731 | 581 | 444 | 315 |
| 037 | 133 | 238 | 348 | 463 | 594 | 728 | 871 | 018 | 172 | 326 | 479 | 621 | 738 | 813 | 825 | 769 | 665 | 523 | 366 | 201 | 039 | 882 | 733 | 591 | 464 | 340 | 229 |
| 967 | 054 | 147 | 246 | 348 | 459 | 572 | 690 | 812 | 933 | 053 | 166 | 266 | 346 | 395 | 400 | 363 | 291 | 193 | 079 | 954 | 827 | 699 | 576 | 458 | 347 | 241 | 140 |

Table 2. Weight table of the 5th neuron of the zero layer for image 1_46 (Fig. 1)

| k = 5, 1_46 | | | | | | | | | | | | | | | | | | | | | | | | | | | |
|---|---|---|---|---|---|---|---|---|---|---|---|---|---|---|---|---|---|---|---|---|---|---|---|---|---|---|---|
| 709 | 795 | 885 | 986 | 092 | 206 | 329 | 456 | 589 | 708 | 803 | 860 | 873 | 849 | 790 | 708 | 614 | 515 | 419 | 324 | 233 | 141 | 052 | 965 | 885 | 811 | 741 | 673 |
| 769 | 862 | 966 | 079 | 206 | 342 | 496 | 664 | 843 | 017 | 161 | 243 | 252 | 199 | 099 | 978 | 848 | 717 | 596 | 478 | 368 | 261 | 157 | 061 | 970 | 885 | 807 | 731 |
| 825 | 929 | 045 | 174 | 321 | 487 | 676 | 894 | 143 | 411 | 650 | 784 | 779 | 657 | 483 | 295 | 114 | 945 | 789 | 645 | 511 | 387 | 269 | 158 | 055 | 961 | 873 | 792 |
| 882 | 993 | 121 | 266 | 434 | 628 | 858 | 138 | 487 | 915 | 354 | 594 | 544 | 264 | 949 | 663 | 415 | 194 | 997 | 823 | 662 | 519 | 384 | 260 | 146 | 038 | 940 | 855 |
| 933 | 053 | 191 | 353 | 538 | 762 | 033 | 376 | 835 | 478 | 248 | 653 | 547 | 007 | 489 | 079 | 745 | 465 | 223 | 012 | 824 | 657 | 504 | 365 | 237 | 117 | 010 | 914 |
| 981 | 107 | 257 | 430 | 634 | 880 | 184 | 579 | 118 | 906 | 881 | 441 | 391 | 779 | 086 | 532 | 099 | 755 | 464 | 213 | 995 | 799 | 627 | 471 | 328 | 197 | 080 | 974 |
| 019 | 156 | 311 | 497 | 716 | 978 | 305 | 727 | 301 | 122 | 162 | 894 | 038 | 560 | 712 | 007 | 479 | 063 | 716 | 421 | 169 | 945 | 751 | 579 | 421 | 278 | 149 | 032 |
| 055 | 197 | 357 | 553 | 778 | 050 | 391 | 819 | 387 | 156 | 137 | 093 | 467 | 144 | 283 | 491 | 877 | 385 | 979 | 638 | 348 | 096 | 877 | 687 | 512 | 358 | 215 | 090 |
| 080 | 228 | 396 | 590 | 824 | 100 | 441 | 868 | 413 | 132 | 093 | 169 | 736 | 608 | 868 | 033 | 314 | 728 | 254 | 860 | 530 | 249 | 002 | 791 | 602 | 432 | 282 | 147 |
| 102 | 250 | 421 | 615 | 851 | 128 | 463 | 879 | 402 | 078 | 992 | 092 | 852 | 010 | 552 | 687 | 812 | 096 | 538 | 086 | 712 | 397 | 127 | 894 | 690 | 507 | 345 | 200 |
| 113 | 263 | 436 | 630 | 862 | 136 | 461 | 861 | 356 | 982 | 795 | 805 | 794 | 252 | 088 | 372 | 334 | 471 | 821 | 311 | 893 | 545 | 248 | 992 | 773 | 575 | 403 | 247 |
| 117 | 266 | 438 | 633 | 858 | 127 | 441 | 822 | 289 | 864 | 593 | 530 | 601 | 307 | 402 | 842 | 785 | 836 | 106 | 535 | 070 | 686 | 364 | 085 | 849 | 641 | 455 | 294 |
| 116 | 263 | 432 | 624 | 842 | 101 | 405 | 767 | 208 | 740 | 404 | 247 | 264 | 218 | 561 | 190 | 226 | 213 | 391 | 754 | 243 | 821 | 473 | 173 | 917 | 698 | 503 | 334 |
| 107 | 251 | 416 | 604 | 816 | 063 | 357 | 700 | 111 | 613 | 228 | 010 | 014 | 103 | 640 | 489 | 720 | 609 | 676 | 967 | 408 | 951 | 571 | 252 | 977 | 745 | 544 | 364 |
| 094 | 234 | 393 | 574 | 780 | 015 | 295 | 618 | 006 | 476 | 054 | 799 | 792 | 942 | 621 | 661 | 052 | 948 | 951 | 177 | 569 | 069 | 659 | 320 | 029 | 786 | 575 | 388 |
| 071 | 208 | 362 | 536 | 734 | 959 | 221 | 528 | 890 | 331 | 871 | 564 | 499 | 638 | 460 | 698 | 279 | 285 | 244 | 389 | 717 | 177 | 735 | 374 | 071 | 814 | 596 | 406 |
| 045 | 176 | 325 | 491 | 680 | 894 | 139 | 428 | 764 | 171 | 669 | 295 | 110 | 130 | 135 | 613 | 450 | 689 | 575 | 608 | 859 | 270 | 797 | 413 | 098 | 831 | 607 | 412 |
| 016 | 140 | 281 | 439 | 618 | 820 | 048 | 318 | 631 | 002 | 454 | 012 | 720 | 635 | 690 | 388 | 485 | 948 | 892 | 831 | 987 | 341 | 836 | 433 | 107 | 832 | 605 | 411 |
| 981 | 099 | 233 | 381 | 551 | 741 | 953 | 200 | 488 | 824 | 232 | 727 | 345 | 137 | 104 | 022 | 371 | 072 | 204 | 033 | 079 | 380 | 849 | 430 | 096 | 820 | 588 | 396 |
| 943 | 054 | 180 | 320 | 476 | 655 | 852 | 075 | 339 | 642 | 005 | 444 | 984 | 676 | 580 | 590 | 110 | 002 | 283 | 114 | 107 | 371 | 824 | 398 | 061 | 791 | 560 | 369 |
| 901 | 008 | 123 | 255 | 401 | 564 | 745 | 950 | 185 | 459 | 780 | 164 | 634 | 237 | 047 | 029 | 674 | 733 | 167 | 090 | 067 | 303 | 751 | 336 | 005 | 740 | 520 | 331 |
| 858 | 958 | 064 | 186 | 322 | 471 | 638 | 822 | 031 | 274 | 554 | 887 | 287 | 791 | 446 | 266 | 040 | 267 | 859 | 946 | 917 | 159 | 631 | 237 | 925 | 674 | 463 | 282 |
| 812 | 907 | 005 | 115 | 240 | 377 | 528 | 694 | 878 | 092 | 332 | 613 | 947 | 351 | 859 | 512 | 245 | 568 | 258 | 410 | 563 | 925 | 460 | 106 | 822 | 590 | 395 | 228 |
| 765 | 852 | 942 | 045 | 159 | 283 | 418 | 566 | 730 | 911 | 116 | 349 | 620 | 935 | 310 | 763 | 266 | 615 | 421 | 752 | 121 | 633 | 253 | 953 | 705 | 495 | 318 | 164 |
| 716 | 796 | 883 | 974 | 077 | 190 | 309 | 442 | 584 | 738 | 912 | 100 | 315 | 554 | 818 | 102 | 379 | 549 | 440 | 075 | 675 | 326 | 031 | 783 | 574 | 391 | 233 | 093 |
| 669 | 741 | 821 | 904 | 997 | 095 | 203 | 320 | 445 | 576 | 719 | 871 | 039 | 215 | 396 | 571 | 711 | 772 | 704 | 514 | 276 | 031 | 810 | 611 | 439 | 282 | 144 | 019 |
| 618 | 686 | 758 | 836 | 918 | 004 | 103 | 202 | 313 | 424 | 538 | 662 | 789 | 922 | 048 | 155 | 230 | 249 | 202 | 089 | 938 | 767 | 602 | 446 | 303 | 173 | 051 | 943 |
| 569 | 631 | 698 | 769 | 840 | 918 | 004 | 092 | 185 | 279 | 374 | 472 | 571 | 669 | 758 | 827 | 870 | 872 | 838 | 765 | 660 | 540 | 413 | 290 | 173 | 065 | 963 | 865 |

Tables 1 and 2 show the weight matrices of the zero layer for the reference images 0_157 and 1_46 from the MNIST test dataset. Each value in the tables corresponds to the total potential at a given cell, induced by the charged surface of a single image.

For the case of a neural network without a zero layer, Table 3 presents an example of a weight matrix for a neuron in the first layer. This matrix is obtained by subtracting the weight matrices of the zero layer (Tables 1 and 2) corresponding to images 0_157 and 1_46. This operation simulates the potential values formed on the plane of potentiometers by the two images shown in Fig. 1.

**Table 3.** Weight matrix of a first-layer neuron performing a comparison of images 0_157 and 1_46

k = 0, k1 = 5, 0_157, 1_46 Wh1 = -26241

| | | | | | | | | | | | | | | | | | | | | | | | | | | | |
|---|---|---|---|---|---|---|---|---|---|---|---|---|---|---|---|---|---|---|---|---|---|---|---|---|---|---|---|
| 18 | 17 | 16 | 04 | 92 | 82 | 60 | 39 | 13 | 8 | 01 | 29 | 88 | 62 | 45 | 18 | 70 | 05 | 18 | 20 | 09 | 97 | 78 | 60 | 42 | 17 | 95 | 78 |
| 27 | 26 | 21 | 12 | 95 | 75 | 46 | 05 | 8 | 6 | 2 | 3 | 22 | 44 | 80 | 87 | 59 | 01 | 10 | 01 | 81 | 57 | 32 | 04 | 76 | 51 | 24 | 01 |
| 41 | 40 | 32 | 21 | 02 | 71 | 30 | 9 | 19 | 116 | 188 | 162 | 14 | 09 | 36 | 07 | 02 | 43 | 37 | 10 | 75 | 30 | 92 | 56 | 19 | 85 | 56 | 27 |
| 56 | 55 | 48 | 37 | 15 | 78 | 24 | 6 | 108 | 316 | 529 | 544 | 284 | 62 | 69 | 30 | 38 | 55 | 16 | 51 | 89 | 22 | 64 | 11 | 63 | 25 | 86 | 47 |
| 74 | 75 | 74 | 59 | 38 | 99 | 39 | 8 | 172 | 530 | 986 | 1063 | 638 | 81 | 91 | 263 | 334 | 275 | 163 | 035 | 22 | 27 | 45 | 71 | 12 | 63 | 19 | 74 |
| 95 | 00 | 97 | 91 | 75 | 43 | 84 | 0 | 145 | 556 | 1095 | 1169 | 626 | 98 | 414 | 910 | 912 | 749 | 489 | 257 | 076 | 40 | 31 | 39 | 65 | 02 | 49 | 01 |
| 18 | 27 | 33 | 31 | 23 | 05 | 63 | 80 | | 319 | 754 | 792 | 267 | 14 | 657 | 273 | 441 | 293 | 858 | 497 | 245 | 067 | 26 | 10 | 18 | 42 | 78 | 30 |
| 42 | 58 | 73 | 76 | 86 | 89 | 75 | 43 | 72 | 46 | 26 | 157 | 18 | 95 | 393 | 027 | 382 | 464 | 141 | 750 | 436 | 204 | 025 | 81 | 71 | 82 | 10 | 51 |
| 72 | 90 | 12 | 38 | 61 | 87 | 13 | 42 | 82 | 55 | 73 | 83 | 81 | 111 | 40 | 122 | 679 | 200 | 322 | 034 | 656 | 352 | 131 | 58 | 24 | 20 | 39 | 73 |
| 96 | 26 | 58 | 98 | 38 | 92 | 63 | 58 | 89 | 072 | 119 | 92 | 350 | 888 | 671 | | 99 | 579 | 177 | 268 | 888 | 515 | 239 | 028 | 70 | 52 | 63 | 91 |
| 29 | 64 | 03 | 58 | 16 | 99 | 18 | 78 | 209 | 469 | 436 | 97 | 711 | 1536 | 1608 | 1022 | 7 | 52 | 777 | 290 | 105 | 688 | 343 | 094 | 08 | 82 | 80 | 05 |
| 56 | 99 | 49 | 15 | 95 | 03 | 63 | 192 | 539 | 911 | 668 | 45 | 893 | 1903 | 2182 | 1708 | 638 | 50 | 457 | 278 | 356 | 869 | 434 | 145 | 38 | 95 | 93 | 10 |
| 80 | 28 | 92 | 65 | 63 | 98 | 088 | 368 | 789 | 201 | 857 | 42 | 782 | 2007 | 2504 | 2187 | 1173 | 1 | 199 | 215 | 440 | 960 | 481 | 165 | 54 | 00 | 93 | 09 |
| 04 | 57 | 25 | 09 | 23 | 77 | 187 | 500 | 959 | 380 | 018 | 93 | 646 | 2001 | 2674 | 2555 | 1710 | 387 | 49 | 092 | 404 | 949 | 478 | 161 | 49 | 92 | 83 | 04 |
| 19 | 79 | 53 | 45 | 68 | 039 | 264 | 601 | 079 | 519 | 166 | 53 | 484 | 1896 | 2698 | 2751 | 2049 | 711 | 06 | 926 | 295 | 881 | 430 | 124 | 21 | 69 | 66 | 90 |
| 34 | 97 | 75 | 70 | 96 | 075 | 319 | 666 | 166 | 632 | 312 | 148 | 228 | 1615 | 2546 | 2781 | 2249 | 1012 | 46 | 729 | 144 | 763 | 340 | 057 | 68 | 35 | 39 | 66 |
| 43 | 09 | 87 | 87 | 16 | 095 | 345 | 702 | 223 | 721 | 457 | 382 | 51 | 1097 | 2189 | 2644 | 2349 | 1335 | 72 | 504 | 936 | 588 | 211 | 65 | 00 | 85 | 05 | 41 |
| 45 | 10 | 92 | 91 | 16 | 092 | 345 | 701 | 229 | 769 | 580 | 627 | 53 | 554 | 1663 | 2308 | 2242 | 1416 | 3 | 243 | 613 | 329 | 040 | 50 | 16 | 31 | 63 | 06 |
| 44 | 11 | 85 | 81 | 02 | 070 | 309 | 650 | 164 | 734 | 648 | 873 | 91 | 2 | 932 | 1764 | 1889 | 1229 | 53 | 43 | 144 | 97 | 37 | 21 | 31 | 69 | 21 | 72 |
| 36 | 00 | 73 | 61 | 78 | 030 | 238 | 544 | 986 | 505 | 584 | 156 | 538 | 51 | 162 | 1068 | 1320 | 810 | 43 | 18 | 37 | 98 | 47 | 95 | 50 | 06 | 74 | 40 |
| 27 | 84 | 56 | 34 | 36 | 70 | 148 | 385 | 700 | 052 | 311 | 440 | 182 | 541 | 56 | 127 | 584 | 428 | 128 | 59 | 54 | 54 | 89 | 83 | 72 | 53 | 30 | 09 |
| 14 | 68 | 28 | 98 | 91 | 01 | 042 | 221 | 439 | 693 | 008 | 370 | 441 | 224 | 756 | 027 | 50 | 138 | 273 | 218 | 0 | 81 | 67 | 99 | 08 | 02 | 91 | 82 |
| 98 | 44 | 00 | 63 | 41 | 32 | 42 | 076 | 233 | 417 | 655 | 919 | 125 | 291 | 161 | 587 | 66 | 54 | 415 | 261 | 5 | 01 | 99 | 42 | 62 | 64 | 59 | 53 |
| 81 | 21 | 70 | 25 | 86 | 63 | 50 | 51 | 063 | 197 | 347 | 508 | 676 | 859 | 928 | 535 | 10 | 49 | 346 | 157 | 6 | 91 | 69 | 07 | 26 | 35 | 34 | 28 |
| 63 | 96 | 37 | 89 | 39 | 95 | 67 | 41 | 24 | 014 | 109 | 211 | 300 | 378 | 372 | 122 | 51 | 01 | 4 | 9 | 35 | 14 | 66 | 93 | 05 | 12 | 15 | 14 |
| 39 | 72 | 05 | 49 | 92 | 39 | 91 | 48 | 05 | 66 | 27 | 85 | 025 | 033 | 82 | 27 | 91 | 64 | 24 | 94 | 10 | 45 | 72 | 87 | 92 | 99 | 00 | 96 |
| 19 | 47 | 80 | 12 | 45 | 90 | 25 | 69 | 05 | 48 | 88 | 17 | 32 | 16 | 65 | 70 | 39 | 16 | 21 | 77 | 63 | 72 | 80 | 87 | 88 | 91 | 89 | 86 |
| 98 | 23 | 49 | 77 | 08 | 41 | 68 | 98 | 27 | 54 | 79 | 94 | 95 | 77 | 37 | 73 | 93 | 19 | 55 | 14 | 94 | 87 | 86 | 86 | 85 | 82 | 78 | 75 |

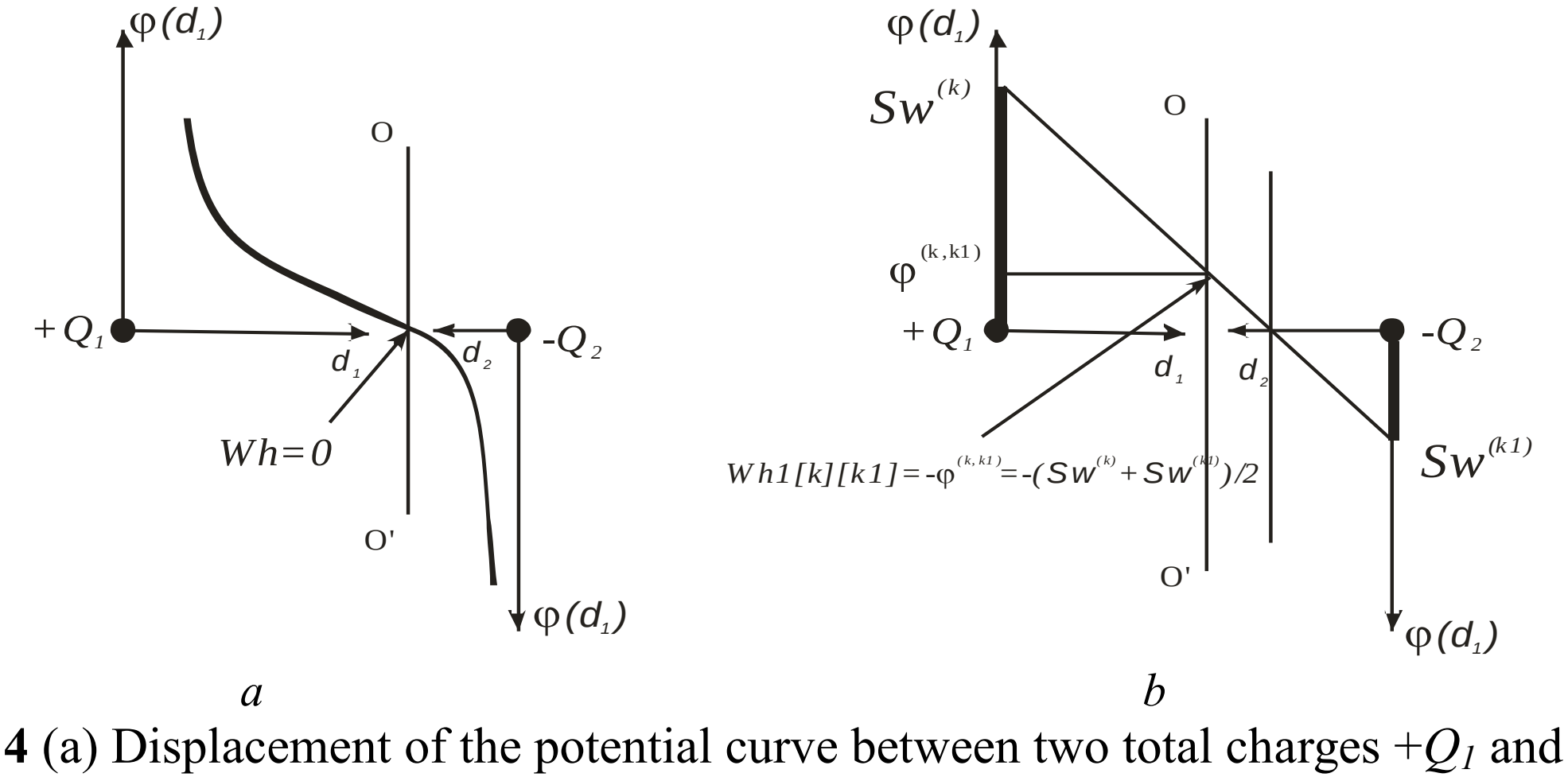

*a* *b*

**Fig. 4** (a) Displacement of the potential curve between two total charges $+Q_1$ and $-Q_2$

(where $|+Q_1| > |-Q_2|$) on two image planes; (b) A method of compensating the displacement by adjusting the neuron's threshold value.

In the previous work [7], potential difficulties associated with the non-uniform distribution and quantity of charge on the two surfaces of the reference images (Fig. 4a) were considered. These issues lead to a shift in the positive and negative potential values relative to the charged regions of the images and the plane of the potentiometer sensors. Possible approaches to address this challenge were also discussed.

In the present example, one possible solution is examined: modifying the threshold value of the first-layer neuron, which also resolves the issue. For this purpose, the threshold value of the first-layer neuron is defined as the average of the state function values of the neuron (2), determined by feeding the reference images themselves — the same ones used to determine the weights for this neuron — into the first-layer neuron.

$$Wh1[k][k1] = -\varphi^{(k,k1)} = -\frac{Sw^{(k)}+Sw^{(k1)}}{2} \text{ , (2)}$$

where 〖$Sw^{(k)}$, $Sw^{(k1)}$ are the state functions of the neurons after feeding the k-th and k1-th reference inputs to the neuron, respectively; Wh1[k][k1] is the weighted threshold value, opposite in sign to the threshold value of the neuron. According to expression (2), all values of the weighted thresholds Wh1[k][k1] for all neurons of the first layer are calculated as shown below:

*k = 0, k1 = 5, n = 157, n1 = 46, Sum = 132216, Sum1 = -79734, Wh1[0][5] = -26241*
*k = 0, k1 = 6, n = 157, n1 = 172, Sum = -81138, Sum1 = -332606, Wh1[0][6] = 206872*
*k = 0, k1 = 7, n = 157, n1 = 35, Sum = 5565, Sum1 = -204329, Wh1[0][7] = 99382*
*k = 0, k1 = 8, n = 157, n1 = 77, Sum = 231886, Sum1 = 57578, Wh1[0][8] = -144732*
*k = 0, k1 = 9, n = 157, n1 = 32, Sum = -2635, Sum1 = -182119, Wh1[0][9] = 92377*
*k = 0, k1 = 10, n = 157, n1 = 51, Sum = -278850, Sum1 = -658880, Wh1[0][10] = 468865*
*k = 0, k1 = 11, n = 157, n1 = 63, Sum = 145446, Sum1 = -25326, Wh1[0][11] = -60060*
*k = 0, k1 = 12, n = 157, n1 = 121, Sum = 137801, Sum1 = 12243, Wh1[0][12] = -75022*
*k = 0, k1 = 13, n = 157, n1 = 24, Sum = 322751, Sum1 = 93421, Wh1[0][13] = -208086*
*k = 0, k1 = 14, n = 157, n1 = 56, Sum = -1337, Sum1 = -185587, Wh1[0][14] = 93462*
...
...
...
*k = 29, k1 = 21, n = 20, n1 = 0, Sum = 406713, Sum1 = 163973, Wh1[29][21] = -285343*
*k = 29, k1 = 22, n = 20, n1 = 64, Sum = 68830, Sum1 = -88930, Wh1[29][22] = 10050*
*k = 29, k1 = 23, n = 20, n1 = 97, Sum = 29550, Sum1 = -131384, Wh1[29][23] = 50917*
*k = 29, k1 = 24, n = 20, n1 = 128, Sum = -206582, Sum1 = -475276, Wh1[29][24] = 340929*
*k = 29, k1 = 25, n = 20, n1 = 61, Sum = 28103, Sum1 = -199785, Wh1[29][25] = 85841*
*k = 29, k1 = 26, n = 20, n1 = 84, Sum = 156810, Sum1 = -9206, Wh1[29][26] = -73802*
*k = 29, k1 = 27, n = 20, n1 = 125, Sum = 347216, Sum1 = 121298, Wh1[29][27] = -234257*
*k = 29, k1 = 28, n = 20, n1 = 151, Sum = 164600, Sum1 = -88706, Wh1[29][28] = -37947*

For example, the value of the threshold weight Wh1[0][5] for a neuron in the first layer, whose weight values are determined based on reference images 0_157 and 1_46, will be:

$$\text{Wh1[0][5]} = -\frac{Sum+Sum1}{2} = -\frac{(132216-79734)}{2} = -26241, \text{(3)}$$

where $Sum = Sw^{(0)}$, $Sum1 = Sw^{(5)}$ in expression (2). This approach is based on the fact that the state function values of a neuron in the first layer should produce the most pronounced response when reference images, which determined the given neuron, are fed to its inputs. For this to happen, the threshold value of the neuron must lie between these two values (see Fig. 4b). The threshold value of a first-layer neuron can be determined as follows. After the neuron's weight table is established (Tables 1 and 2), the two reference patterns used to determine the neuron's weights are fed to the neuron sequentially. Based on the resulting neuron state function values, the neuron's threshold value is calculated. The calculation of the average value can also be performed by neurons, as shown in Fig. 5.

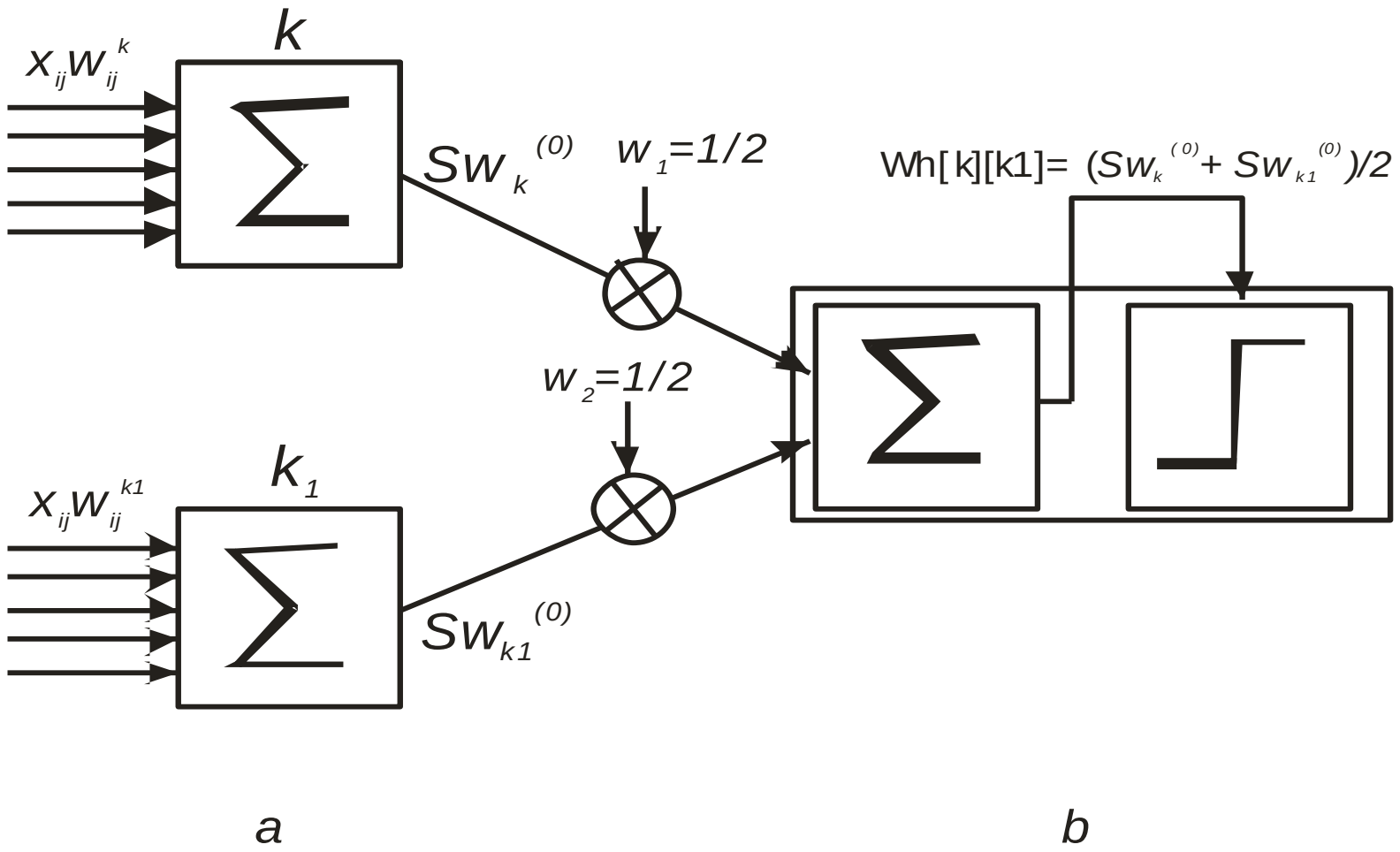

**Fig. 5.** Diagram for determining the weight threshold value Wh1[k][k1] in the neuron *(k, $k_1$)* of the first layer (b), based on the state function values $Sw_k^{(0)}$, $Sw_{k1}^{(0)}$ of neurons *k* and *k1* in the zero layer (a).

**Fig. 6.** A set of reference samples consisting of 30 handwritten digit images selected from the MNIST test dataset. The label of each image is shown above it.

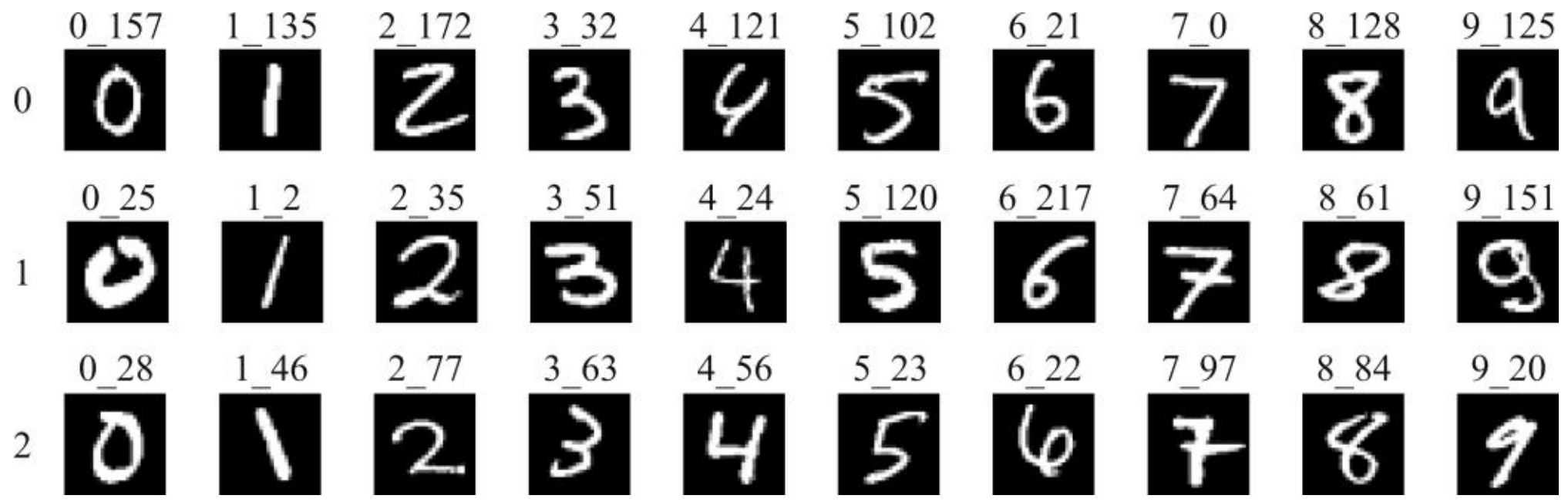

Fig. 6. Thirty selected reference images

In this set of reference images, the weight tables (similar to Table 3) of the first layer of the neural network are determined. The number of these tables for the extended architecture of the neural network based on metric recognition methods (see Fig. 2) is defined by the following expression:

$$n = N(N-1) = 30 * 29 = 870, \quad (4)$$

where *N* is the number of reference patterns

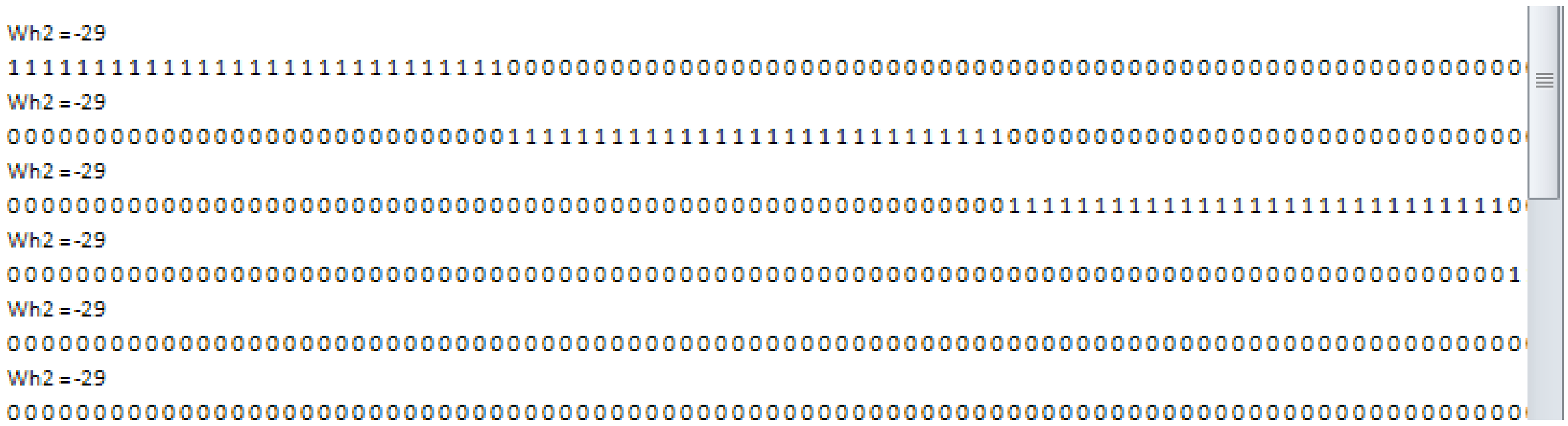

(a)

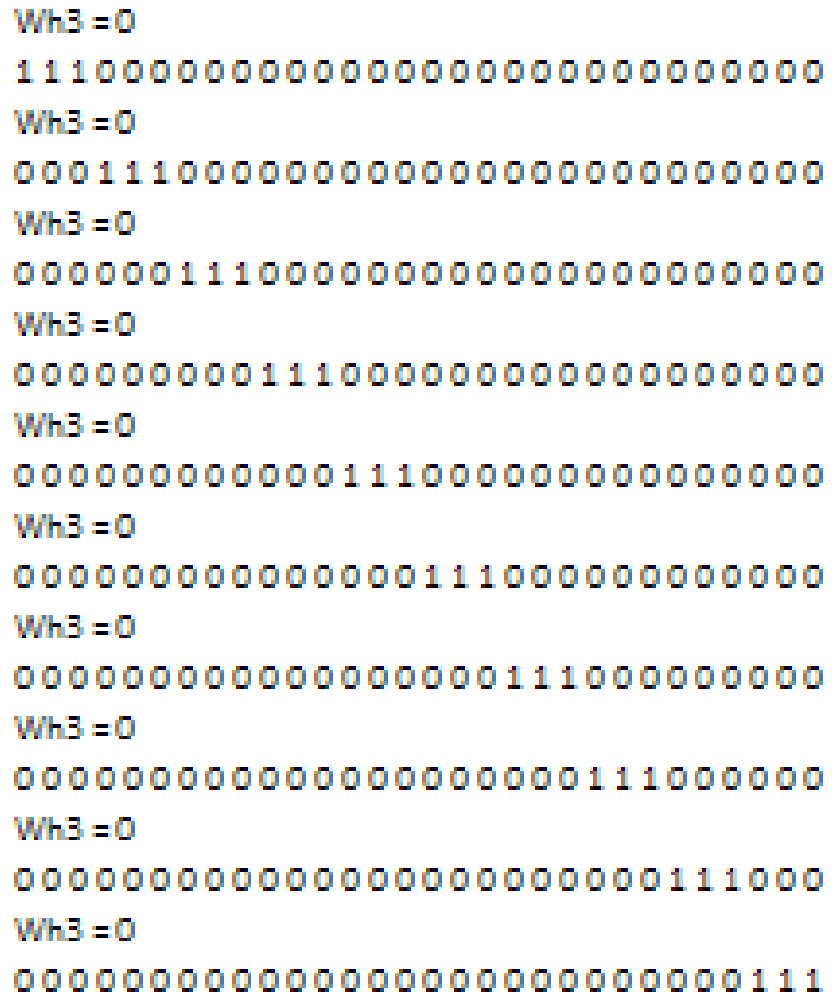

(б)

Fig. 7. Weight values of the second and third layers for 30 reference patterns and 10 input images.(a) Fragment of the weights for the second layer, (b) All weights of the third layer.

The weight values of the second and third layers take values of 1 and 0, where the zero weight values are assigned to the additional connections [5] of the neurons in the second and third layers when transforming the neural network from Fig. 2 into a fully connected neural network. Figure 7 shows fragments of the weight values of the second and third layers. Threshold activation functions are used in the neural network.For the network diagram in Fig. 1, the state function $Sn_{i,j}^{(1)}$ and the activation function $f\left(Sn_{i,j}^{(1)}\right)$ for each neuron in the first layer are defined by the following formulas:

$$Sn_{i,j}^{(1)} = \sum_{r=0}^{R} \sum_{c=0}^{C} x_{c,r} w_{c,r}^{(1)} \quad , \qquad (4)$$

$$\begin{cases} f\left(Sn_{i,j}^{(1)}\right) = 1, \text{ если } Sn_{i,j}^{(1)} < 0 \\ f\left(Sn_{i,j}^{(1)}\right) = 0, \text{ если } Sn_{i,j}^{(1)} > 0 \end{cases} , \qquad (5)$$

where C and R are the number of columns and rows in the weight table (for MNIST, C=28, R=28).

The state functions ($Sw_k^{(2)}$) and activation functions $f\left(Sw_k^{(2)}\right)$of the neurons in the second layer are defined by the following expressions:

$$Sw_k^{(2)} = \sum_{j=1, j \neq k}^{N} f\left(Sw_{k,j}^{(1)}\right), \qquad (7)$$

$$\begin{cases} f\left(Sw_k^{(2)}\right) = 1, \text{если } Sw_k^{(2)} \geq (N-1) = B^{(2)} \\ f\left(Sw_k^{(2)}\right) = 0, \text{ если } Sw_k^{(2)} < (N-1) = B^{(2)} \end{cases}. \quad (8)$$

Here, $B^{(2)}$ = $-Wh1[k][k1] = N\text{-}1$ is the threshold value of the second layer neuron, where NNN is the number of prototypes. The state functions $Sw_k^{(3)}$ and the activation function $f\left(Sw_k^{(3)}\right)$ of the neurons in the third layer are determined by the following expressions:

$$Sw_k^{(3)} = \sum_{i \in k}^{K_k} f\left(Sw_i^{(2)}\right), \quad (9)$$

$$\begin{cases} f\left(Sw_k^{(3)}\right) = 1, \text{ если } Sw_k^{(3)} > 0 \\ f\left(Sw_k^{(3)}\right) = 0, \text{ если } Sw_k^{(3)} \leq 0 \end{cases}, \quad (10)$$

where $K_k$ - is the number of reference patterns for the k-th recognized image.

## 3. Testing the created neural network on the MNIST validation set

The performance of the created neural network is also tested in the software module implemented in the Builder C++ environment, as shown in Fig. 8. A three-layer neural network, depicted in Fig. 2 (with the zero layer not being used), is employed.

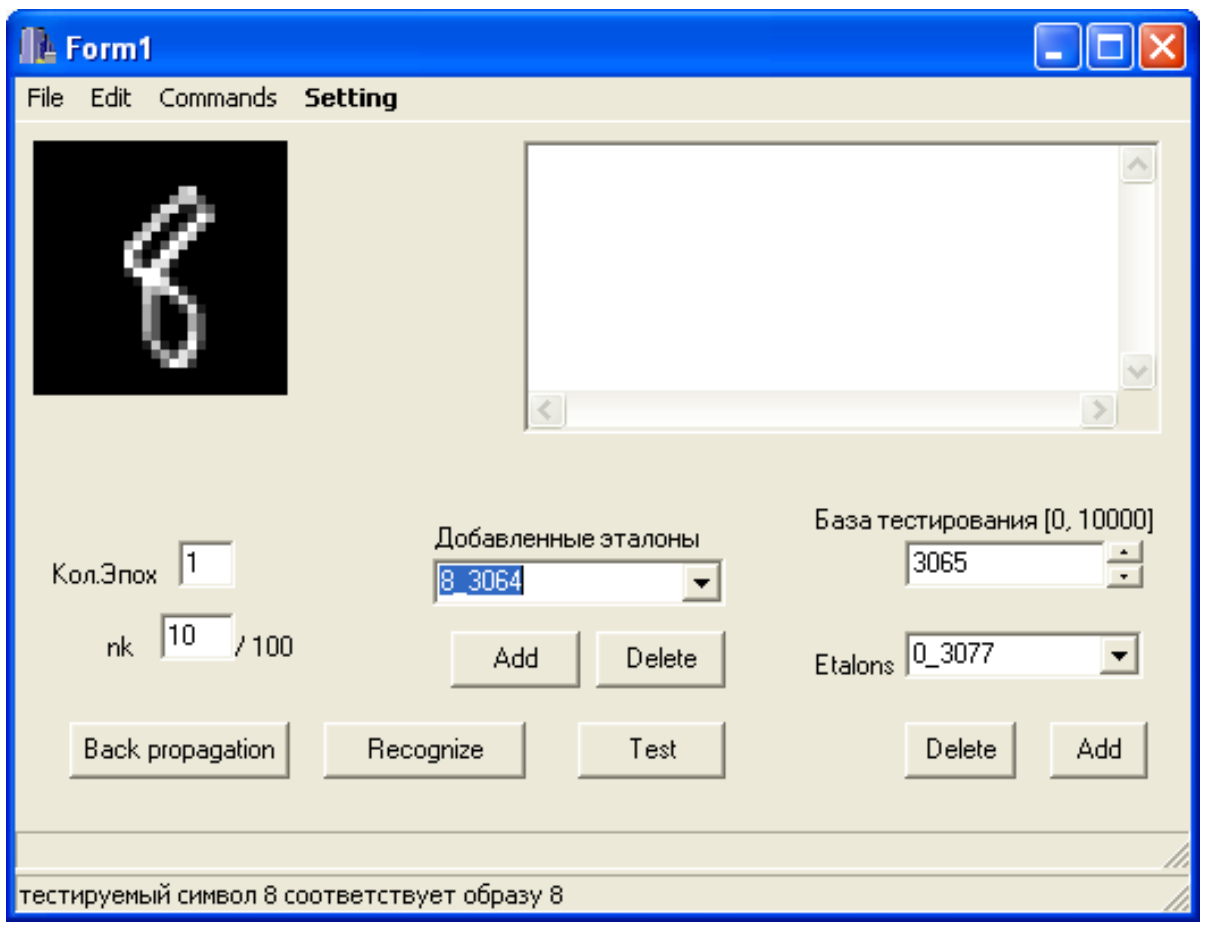

**Fig. 8** Program module implementing the creation of a neural network and testing the network based on MNIST.

Table 4 presents the recognition results on the MNIST test set using 30 reference patterns shown in Fig. 6. The table provides results both separately for each digit image (ij - total number of digit images of pattern j in the MNIST test set, sj - number of correctly identified images of pattern j), as well as the overall result, which amounted to 5047 correctly identified images (50%).

**Table 4** Results of neural network testing on the MNIST test set with initial conditions: N = 30 reference patterns, $q = 10^{-9}$ C, $d_2 = 4$ cm, $d_1 = 2$ cm.

| | | |
|---|---|---|
| s0 = 701 | i0 = 980 | p0 = 71% |
| s1 = 936 | i1 = 1135 | p1 = 82% |
| s2 = 296 | i2 = 1032 | p2 = 28% |
| s3 = 361 | i3 = 1010 | p3 = 35% |
| s4 = 434 | i4 = 982 | p4 = 44% |
| s5 = 357 | i5 = 892 | p5 = 40% |
| s6 = 413 | i6 = 958 | p6 = 43% |
| s7 = 504 | i7 = 1028 | p7 = 49% |
| s8 = 500 | i8 = 974 | p8 = 51% |
| s9 = 545 | i9 = 1009 | p9 = 54% |
| 10000, 5047, 50% | | |

Table 5. Results of testing the neural network on the MNIST test set with initial conditions: N = 30 reference samples, $q = 10^{-9}$ C, $d_2 = 2$ cm, $d_1 = 2$ cm.

| | | |
|---|---|---|
| s0 = 729 | i0 = 980 | p0 = 74% |
| s1 = 987 | i1 = 1135 | p1 = 86% |
| s2 = 307 | i2 = 1032 | p2 = 29% |
| s3 = 387 | i3 = 1010 | p3 = 38% |
| s4 = 468 | i4 = 982 | p4 = 47% |
| s5 = 385 | i5 = 892 | p5 = 43% |
| s6 = 484 | i6 = 958 | p6 = 50% |
| s7 = 532 | i7 = 1028 | p7 = 51% |
| s8 = 477 | i8 = 974 | p8 = 48% |
| s9 = 608 | i9 = 1009 | p9 = 60% |
| | | |
| 10000, 5364, 53% | | |

Table 5 presents the results of recognizing the MNIST test database after reducing the distance between the two image planes (Fig. 1) to 4 cm and, accordingly, to the plane of the sensors, $d_2 = 2$ cm. According to the results in Table 5, it can be seen that the performance improved by 3%, which indicates that changes in the initial physical parameters of the simulated system, such as the value of the point charge q, the distance between the weight table cells d1 (the distance between the sensors), and the distance between the image plane and the potentiometer sensor plane d2, also affect the recognition results.

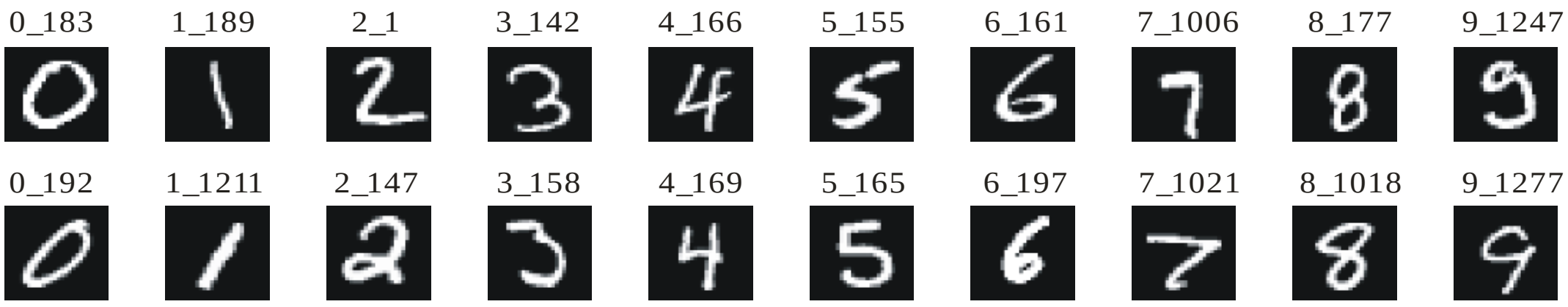

**Fig. 9** Added reference set for the experiment with 50 references.

**Table 6** Testing results on the MNIST benchmark using 50 references.

| | | |
|---|---|---|
| s0 = 852 | i0 = 980 | p0 = 87% |
| s1 = 1123 | i1 = 1135 | p1 = 99% |
| s2 = 516 | i2 = 1032 | p2 = 50% |
| s3 = 585 | i3 = 1010 | p3 = 58% |
| s4 = 775 | i4 = 982 | p4 = 79% |
| s5 = 588 | i5 = 892 | p5 = 66% |
| s6 = 555 | i6 = 958 | p6 = 58% |
| s7 = 616 | i7 = 1028 | p7 = 60% |
| s8 = 633 | i8 = 974 | p8 = 65% |
| s9 = 726 | i9 = 1009 | p9 = 72% |
| | | |
| 10000, 6969, 70% | | |

Table 6 presents the results of recognition on the MNIST test set after adding reference patterns (Fig. 6), with two additional reference patterns for each image, as shown in Fig. 9. Similar to the reference base in Fig. 6, the selection of reference patterns in Fig. 9 was done intuitively and randomly. After adding the new reference patterns, new neurons are added to the first layer of the neural network in a cascade manner, and their weight tables are calculated. The weight tables of the other neurons remain unchanged. The testing results on the MNIST test set with the new reference base increased the performance to 70% (Table 6).

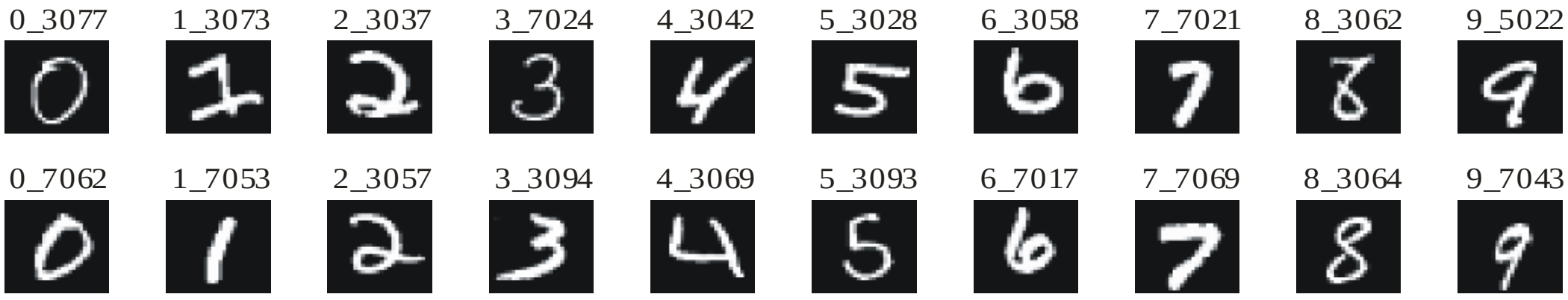

Fig. 10. 20 additional reference samples added for the experiment with 70 reference samples.

Table 7. Test results on the MNIST control dataset using 70 reference samples.

| | | |
|---|---|---|
| s0 = 882 | i0 = 980 | p0 = 90% |
| s1 = 1123 | i1 = 1135 | p1 = 99% |
| s2 = 651 | i2 = 1032 | p2 = 63% |
| s3 = 711 | i3 = 1010 | p3 = 70% |
| s4 = 785 | i4 = 982 | p4 = 80% |
| s5 = 633 | i5 = 892 | p5 = 71% |
| s6 = 689 | i6 = 958 | p6 = 72% |
| s7 = 668 | i7 = 1028 | p7 = 65% |
| s8 = 742 | i8 = 974 | p8 = 76% |
| s9 = 837 | i9 = 1009 | p9 = 83% |
| | | |
| 10000, 7721, 77% | | |

В таблице 7 приведен результат распознавания контрольной базы MNIST после добавления в базу эталонов еще 20 эталонов на рис.10, увеличив общее количество эталонов до 70 эталонов (по 7 эталонов от каждого образа). Результаты распознавания контрольной базы MNIST увеличили количество правильно идентифицированных изображений до 7721 изображений (77%).

Table 7 shows the recognition results of the MNIST test set after adding 20 additional reference patterns, as shown in Figure 10, increasing the total number of reference patterns to 70 (7 reference patterns for each image). The recognition results for the MNIST test set increased the number of correctly identified images to 7,721 images (77%).

From the results in the tables, it can be observed that increasing the number of reference patterns for a single image also increases the percentage of correctly recognized images. Moreover, adding reference patterns to the neural network based on metric recognition methods does not change the previous weight values of the network [4], and the number of neurons increases in a cascading manner. This feature also allows for adding reference patterns for new images to the neural network in the same way [4]. When new reference patterns are added, the weight values are determined only for the newly added neurons.

**Conclusion**. Thus, the experiments conducted above showed that a neural network with weight values determined based on the characteristics of the electrostatic field is feasible, suggesting that the neural network is capable of instantly memorizing and recognizing objects and images without the need for lengthy training and training samples. The resulting network works, and as the number of references increases, the number of recognizable objects also grows. At the same time, the majority of the MNIST database is recognized, and the number of selected references is significantly smaller compared to the number of recognized objects in the recognition task. The references themselves can be selected either randomly, intuitively, or by using additional algorithms (3) that determine the importance of a given reference in the process of the neural network's operation and object recognition. In this process, less significant and rarely used references can, on the contrary, be excluded from the neural network. In biological terms, the neurons corresponding to specific references may either die or survive during the operation of the system. Similarly, references from new images can also be added to the neural network, expanding the set of recognizable objects in the task. The capabilities of neural networks based on metric recognition methods allow this to be done quite easily, without changing the previously set weight values of the neural network. The implemented neural network system with determined weight values can be further trained using existing feedforward neural network training algorithms, which indicates that the ability to predefine and instantly set weight values does not exclude the main capability of the neural network — its ability to learn.